\documentclass[11pt,a4paper]{article}
\usepackage[hyperref]{emnlp-ijcnlp-2019}
\usepackage[T1]{fontenc}
\usepackage{times}
\usepackage{latexsym}

\usepackage{url}
\usepackage{multirow}
\usepackage{todonotes}
\usepackage{soul}

\definecolor{frenchblue}{rgb}{0.0, 0.45, 0.73}
\definecolor{lasallegreen}{rgb}{0.03, 0.47, 0.19}
\definecolor{raspberry}{rgb}{0.89, 0.04, 0.36}
\definecolor{goldenpoppy}{rgb}{0.6, 0.4, 0.08}

\usepackage{scalerel}
\usepackage{tikz}
\usetikzlibrary{svg.path}

\definecolor{orcidlogocol}{HTML}{A6CE39}
\tikzset{
  orcidlogo/.pic={
    \fill[orcidlogocol] svg{M256,128c0,70.7-57.3,128-128,128C57.3,256,0,198.7,0,128C0,57.3,57.3,0,128,0C198.7,0,256,57.3,256,128z};
    \fill[white] svg{M86.3,186.2H70.9V79.1h15.4v48.4V186.2z}
                 svg{M108.9,79.1h41.6c39.6,0,57,28.3,57,53.6c0,27.5-21.5,53.6-56.8,53.6h-41.8V79.1z M124.3,172.4h24.5c34.9,0,42.9-26.5,42.9-39.7c0-21.5-13.7-39.7-43.7-39.7h-23.7V172.4z}
                 svg{M88.7,56.8c0,5.5-4.5,10.1-10.1,10.1c-5.6,0-10.1-4.6-10.1-10.1c0-5.6,4.5-10.1,10.1-10.1C84.2,46.7,88.7,51.3,88.7,56.8z};
  }
}

\newcommand\orcidicon[1]{\href{https://orcid.org/#1}{\mbox{\scalerel*{
\begin{tikzpicture}[yscale=-1,transform shape]
\pic{orcidlogo};
\end{tikzpicture}
}{|}}}}

\usepackage{hyperref}

\aclfinalcopy 

\definecolor{orcidlogocol}{HTML}{A6CE39}

\title{SAMSum Corpus: A Human-annotated Dialogue Dataset\\ for Abstractive Summarization}

\author{Bogdan Gliwa, Iwona Mochol, Maciej Biesek, Aleksander Wawer \orcidicon{0000-0002-7081-9797} \\
  Samsung R\&D Institute Poland \\
  {\tt \{b.gliwa, i.mochol, m.biesek, a.wawer\}@samsung.com} \\}

\date{September 2019}

\begin{document}
\maketitle
\begin{abstract}
This paper introduces the \textbf{SAMSum Corpus}, a~new dataset with abstractive dialogue summaries. We investigate the challenges it poses for automated summarization by testing several  models and comparing their results with those obtained on a~corpus of news articles. We show that model-generated summaries of dialogues achieve higher ROUGE scores than the model-generated summaries of news -- in contrast with human evaluators' judgement. This suggests 
that a~challenging task of abstractive dialogue summarization requires dedicated models and non-standard quality measures. To our knowledge, our study is the first attempt to introduce a high-quality chat-dialogues corpus, manually annotated with abstractive summarizations, which can be used by the research community for further studies.
\end{abstract}

\section{Introduction and related work}

The goal of the summarization task is condensing a piece of text into a shorter version that covers the main points succinctly. 
In the abstractive approach important pieces of information are presented using words and phrases not necessarily appearing in the source text. This requires natural language generation techniques with high level of semantic understanding \cite{abstractive-sum1, abstractive-sum2, single-trans,bert-sum, pointergenerator,fastabs,bottom-up}.


Major research efforts have focused so far on summarization of single-speaker documents like news (e.g., \citet{rnn-abs-sum}) or scientific publications (e.g., \citet{sum-papers}). 
One of the reasons is the availability of large, high-quality news datasets with annotated summaries, e.g., CNN/Daily Mail \cite{cnn-dm, rnn-abs-sum}. 
Such a comprehensive dataset for dialogues is lacking. 

The challenges posed by the abstractive dialogue summarization task have been discussed in the literature with regard to AMI meeting corpus \cite{ami}, e.g. \citet{dials-dep-graph}, \citet{sum-queries}, \citet{dial-sum}. 
Since the corpus has a low number of summaries (for 141 dialogues), \citet{dial-sum} proposed to use assigned topic descriptions as gold references. These are short, label-like goals of the meeting, e.g., \textit{costing evaluation of project process; components, materials and energy sources; chitchat}. Such descriptions, however, are very general, lacking the messenger-like structure and any information about the speakers. 

To benefit from large news corpora, \citet{sum-2019} built a dialogue summarization model that first converts a conversation  into  a  structured  text document and later applies an attention-based pointer network to create an abstractive summary. Their model, trained on structured text documents of CNN/Daily Mail dataset, was evaluated on the Argumentative Dialogue Summary Corpus \cite{sum-argum-corpus}, which, however, contains only 45 dialogues.

In the present paper, we further investigate the problem of abstractive dialogue summarization. 
With the growing popularity of online conversations via applications like Messenger, WhatsApp and WeChat, summarization of chats between a~few participants is a~new interesting direction of summarization research. 
For this purpose we have created the \textbf{SAMSum Corpus}\footnote{The name is a shortcut for \textbf{S}amsung \textbf{A}bstractive \textbf{M}essenger \textbf{Sum}marization} which contains over 16k chat dialogues with manually annotated summaries. The dataset is freely available for the research community\footnote{The dataset is shared on terms of the Attribution-NonCommercial-NoDerivatives 4.0 International (CC BY-NC-ND 4.0) license. It accompanies this paper on arXiv.}.

The paper is structured as follows: in Section~\ref{Dataset} we present details about the new corpus and describe how it was created, validated and cleaned. Brief description of baselines used in the summarization task can be found in Section~\ref{Baselines}. In Section~\ref{Experimental-setup}, we describe our experimental setup and parameters of models. Both evaluations of summarization models, the automatic with ROUGE metric and the linguistic one, are reported in Section~\ref{Results} and Section~\ref{Linguistic}, respectively. Examples of models' outputs and some errors they make are described in Section~\ref{Difficulties}. Finally, discussion, conclusions and ideas for further research are presented in sections \ref{Discussion} and \ref{Conclusion}.

\section{SAMSum Corpus} \label{Dataset}
\begin{table}[t!]
\begin{center}
\begin{tabular}{|l|c c c|}
\hline \bf Dataset & \bf Train & \bf Validation & \bf Test \\ \hline
CNN/DM & 287 227 & 13 368 & 11 490 \\
SAMSum & 14 732 & 818 & 819 \\
\hline
\end{tabular}
\end{center}
\caption{Datasets sizes}
\label{table:datasets-split}
\end{table}




{\bf Initial approach.} Since there was no available corpus of messenger conversations, we considered two approaches to build it: (1) using existing datasets of documents, which have a form similar to chat conversations, (2) creating such a dataset by linguists.

In the first approach, we reviewed datasets from the following categories: chatbot dialogues, SMS corpora, IRC/chat data, movie dialogues, tweets, comments data (conversations formed by replies to comments), transcription of meetings, written discussions, phone dialogues and daily communication data. Unfortunately, they all differed in some respect from the conversations that are typically written in messenger apps, e.g. they were too technical (IRC data), too long (comments data, transcription of meetings), lacked context (movie dialogues) or they were more of a spoken type, such as a dialogue between a petrol station assistant and a~client buying petrol. 

As a~consequence, we decided to create a chat dialogue dataset by constructing such conversations that would epitomize the style of a messenger app.

{\bf Process of building the dataset.} Our dialogue summarization dataset contains natural messenger-like conversations created and written down by linguists fluent in English. The style and register of conversations are diversified -- dialogues could be informal, semi-formal or formal, they may contain slang phrases, emoticons and typos. We asked linguists to create conversations similar to those they write on a~daily basis, reflecting the proportion of topics of their real-life messenger conversations. It includes chit-chats, gossiping about friends, arranging meetings, discussing politics, consulting university assignments with colleagues, etc.  Therefore, this dataset does not contain any sensitive data or fragments of other corpora. 

Each dialogue was created by one person. After collecting all of the conversations, we asked language experts to annotate them with summaries, assuming that they should (1) be rather short, (2) extract important pieces of information, (3) include names of interlocutors, (4) be written in the third person. Each dialogue contains only one reference summary. 
 
{\bf Validation.} Since the SAMSum corpus contains dialogues created by linguists, the question arises whether such conversations are really similar to those typically written via messenger apps. To find the answer, we performed a validation task. We asked two linguists to doubly annotate 50  conversations in order to verify whether the dialogues could appear in a messenger app and could be summarized (i.e. a dialogue is not too general or unintelligible) or not (e.g. a dialogue between two people in a shop). The results revealed that 94\% of examined dialogues were classified by both annotators as good i.e. they do look like conversations from a messenger app and could be condensed in a~reasonable way.
In a similar validation task, conducted for the existing dialogue-type datasets (described in the Initial approach section), the annotators agreed that only 28\% of the dialogues resembled conversations from a messenger app. 

{\bf Cleaning data.} After preparing the dataset, we conducted a process of cleaning it in a semi-automatic way. Beforehand, we specified a format for written dialogues with summaries: a colon should separate an author of utterance from its content, each utterance is expected to be in a separate line.
Therefore, we could easily find all deviations from the agreed structure -- some of them could be automatically fixed (e.g. when instead of a colon, someone used a semicolon right after the interlocutor's name at the beginning of an utterance), others were passed for verification to linguists. We also tried to correct typos in interlocutors' names (if one person has several utterances, it happens that, before one of them, there is a typo in his/her name) -- we used the Levenshtein distance to find very similar names (possibly with typos e.g. 'George' and 'Goerge') in a single conversation, and those cases with very similar names were passed to linguists for verification.

{\bf Description.} The created dataset is made of 16369 conversations distributed uniformly into 4~groups based on the number of utterances in conversations: 3-6, 7-12, 13-18 and 19-30. Each utterance contains the name of the speaker. Most conversations consist of dialogues between two interlocutors (about 75\% of all conversations), the rest is between three or more people. Table~\ref{table:datasets-split} presents the size of the dataset split used in our experiments. The example of a~dialogue from this corpus is shown in Table \ref{table:dialog-example}.



\begin{table}[ht!]
\begin{center}
\begin{tabular}{|l|}
\hline \bf Dialogue \\ \hline
\textcolor{lasallegreen}{Blair:} Remember we are seeing the wedding\\ planner after work \\
\textcolor{raspberry}{Chuck:} Sure, where are we meeting her? \\
\textcolor{lasallegreen}{Blair:} At Nonna Rita's \\
\textcolor{raspberry}{Chuck:} Can I order their seafood tagliatelle\\ or are we just having coffee with her? I've\\ been dreaming about it since we went there\\ last month \\
\textcolor{lasallegreen}{Blair:} Haha sure why not \\
\textcolor{raspberry}{Chuck:} Well we both remmber the spaghetti\\ pomodoro disaster from our last meeting with\\ Diane \\
\textcolor{lasallegreen}{Blair:} Omg hahaha it was all over her white\\ blouse \\
\textcolor{raspberry}{Chuck:} :D \\
\textcolor{lasallegreen}{Blair:} :P \\
\hline \bf Summary \\ \hline
Blair and Chuck are going to meet the\\ wedding planner after work at Nonna Rita's.\\ The tagliatelle served at Nonna Rita's are\\ very good. \\
\hline
\end{tabular}
\end{center}
\caption{Example of a dialogue from the collected corpus}
\label{table:dialog-example}
\end{table}

\section{Dialogues baselines} \label{Baselines}
The baseline commonly used in the news summarization task is Lead-3  \cite{pointergenerator}, which takes three leading sentences of the document as the summary. The underlying assumption is that the beginning of the article contains the most significant information. Inspired by the Lead-n model, we propose a few different simple models:
\begin{itemize}
    \item MIDDLE-n, which takes n~utterances from the middle of the dialogue,
    \item LONGEST-n, treating only n~longest utterances in order of length as a~summary,
    \item LONGER-THAN-n, taking only utterances longer than n~characters in order of length (if there is no such long utterance in the dialogue, takes the longest one),
    \item MOST-ACTIVE-PERSON, which treats all utterances of the most active person in the dialogue as a~summary.
\end{itemize}
Results of the evaluation of the above models are reported in Table~\ref{table:dialogues-baselines}. There is no obvious baseline for the task of dialogues summarization. We expected rather low results for Lead-3, as the beginnings of the conversations usually contain greetings, not the main part of the discourse. 
However, it seems that in our dataset greetings are frequently combined with question-asking or information passing (sometimes they are even omitted) and such a baseline works even better than the MIDDLE baseline (taking utterances from the middle of a dialogue). Nevertheless, the best dialogue baseline turns out to be the LONGEST-3 model.


\begin{table}[t!]
\begin{center}
\begin{tabular}{|l|r|c c c|}
\hline \bf Model & \bf n & \bf R-1 & \bf R-2 & \bf R-L \\ \hline
\multirow{3}{*}{LEAD}           & 3 & 31.40 & 8.68 & 29.42 \\
                                & 4 & 31.87 & 8.93 & 29.91 \\
                                & 5 & 32.02 & 9.53 & \textbf{30.07} \\
\hline
\multirow{3}{*}{MIDDLE}         & 3 & 28.04 & 6.57 & 26.13 \\
                                & 4 & 30.08 & 7.96 & 28.10 \\
                                & 5 & 29.91 & 8.12 & 27.97 \\
\hline
\multirow{3}{*}{LONGEST}        & 3 & \textbf{32.46} & 10.27 & 29.92 \\
                                & 4 & 32.19 & \textbf{10.35} & 29.91 \\
                                & 5 & 31.61 & 10.21 & 29.55 \\
\hline
\multirow{3}{*}{\shortstack[l]{LONGER\\-THAN}} & 10 & 28.31 &  9.69 & 26.72 \\                                                      & 20 & 29.36 & 10.23 & 27.59 \\
                                               & 30 & 29.61 & 10.28 & 27.71 \\
\hline
MOST-ACTIVE                     & \multirow{2}{*}{n/a} & \multirow{2}{*}{26.54}
                                & \multirow{2}{*}{8.55} & \multirow{2}{*}{24.57} \\
-PERSON                         &   &   &   &  \\
\hline
\end{tabular}
\end{center}
\caption{Baselines for the dialogues summarization}
\label{table:dialogues-baselines}
\end{table}

\section{Experimental setup} \label{Experimental-setup}
This section contains a description of setting used in the experiments carried out.

\subsection{Data preparation}
In order to build a dialogue summarization model, we adopt the following strategies: (1)~each candidate architecture is trained and evaluated on the dialogue dataset; (2)~each architecture is trained on the train set of CNN/Daily Mail joined together with the train set of the dialogue data, and evaluated on the dialogue test set. 

In addition, we prepare a version of dialogue data, in which utterances are separated with a special token called the separator (artificially added token e.g. '$<$EOU$>$' for models using word embeddings, '$|$' for models using subword embeddings). 
In all our experiments, news and dialogues are truncated to 400 tokens, and summaries -- to 100 tokens. The maximum length of generated summaries was not limited. 

\subsection{Models} \label{models-params}
We carry out experiments with the following summarization models (for all architectures we set the beam size for beam search decoding to 5):
\begin{itemize}
    \item \textbf{Pointer generator network} \cite{pointergenerator}.
    In the case of {\it Pointer Generator}, we use a~default configuration\footnote{\url{https://github.com/abisee/pointer-generator}}, changing only the minimum length of the generated summary from 35 (used in news) to 15 (used in dialogues).
    
    \item \textbf{Transformer} \cite{transformer}.
    The model is trained using OpenNMT library\footnote{\url{https://github.com/OpenNMT/OpenNMT-py}}. We use the same parameters for training both on news and on dialogues\footnote{\url{http://opennmt.net/OpenNMT-py/Summarization.html}}, changing only the minimum length of the generated summary -- 35 for news and 15 for dialogues.
    
    \item \textbf{Fast Abs RL} \cite{fastabs}.
    It is trained using its default parameters\footnote{\url{https://github.com/ChenRocks/fast_abs_rl}}. For dialogues, we change the convolutional word-level sentence encoder (used in extractor part) to only use kernel with size equal 3~instead of 3-5 range. It is caused by the fact that some of utterances are very short and the default setting is unable to handle that.
    
    \item \textbf{Fast Abs RL Enhanced}.
    The additional variant of the \textit{Fast Abs RL} model with slightly changed utterances i.e. to each utterance, at the end, after artificial separator, we add names of all other interlocutors. The reason for that is that \textit{Fast Abs RL} requires text to be split into sentences (as it selects sentences and then paraphrase each of them). For dialogues, we divide text into utterances (which is a~natural unit in conversations), so sometimes, a single utterance may contain more than one sentence. Taking into account how this model works, it may happen that it selects an utterance of a single person (each utterance starts with the name of the author of the utterance) and has no information about other interlocutors (if names of other interlocutors do not appear in selected utterances), so it may have no chance to use the right people's names in generated summaries.
    
    \item \textbf{LightConv and DynamicConv} \cite{lightconv}.
    The implementation is available in fairseq\footnote{\url{https://github.com/pytorch/fairseq}} \cite{fairseq}.  We train lightweight convolution models in two manners: (1)~learning token representations from scratch; in this case we apply BPE tokenization with the vocabulary of 30K types, using fastBPE implementation\footnote{\url{https://github.com/glample/fastBPE}} \cite{fastBPE}; (2)~initializing token embeddings with pre-trained language model representations; as a~language model we choose GPT-2 small \cite{gpt-2}. 
\end{itemize}

\subsection{Evaluation metrics}
We evaluate models with the standard ROUGE metric~\cite{lin-2004-rouge}, reporting the $F_1$ scores (with stemming) for ROUGE-1, ROUGE-2 and ROUGE-L following previous works~\cite{fastabs, pointergenerator}. We obtain scores using the \texttt{py-rouge} package\footnote{\url{https://pypi.org/project/py-rouge/}}. 



\section{Results} \label{Results}
The results for the news summarization task are shown in Table~\ref{table:news-results} and for the dialogue summarization -- in Table~\ref{table:dialogues-results}. 
In both domains, the best models' ROUGE-1 exceeds $39$, ROUGE-2 -- $17$ and ROUGE-L -- $36$. 
Note that the strong baseline for news (Lead-3) is outperformed in all three metrics only by one model. In the case of dialogues, all tested models perform better than the baseline (LONGEST-3). 

In general, the Transformer-based architectures benefit from training on the joint dataset: news+dialogues, even though the news and the dialogue documents have very different structures. Interestingly, this does not seem to be the case for the {\it Pointer Generator} or {\it Fast Abs RL} model. 

The inclusion of a separation token between dialogue utterances is advantageous for most models -- presumably because it improves the discourse structure. The improvement is most visible when training is performed on the joint dataset.

Having compared two variants of the {\it Fast Abs RL} model -- with original utterances and with enhanced ones (see Section \ref{models-params}), we conclude that enhancing utterances with information about the other interlocutors helps achieve higher ROUGE values. 

The largest improvement of the model performance is observed for {\it LightConv} and {\it DynamicConv} models when they are complemented with pretrained embeddings from the language model {\it GPT-2}, trained on enormous corpora.

It is also worth noting that some models ({\it Pointer Generator}, {\it Fast Abs RL}), trained only on the dialogues corpus (which has 16k dialogues), reach similar level (or better) in terms of ROUGE metrics than models trained on the CNN/DM news dataset (which has more than 300k articles). Adding pretrained embeddings and training on the joined dataset helps in achieving significantly higher values of ROUGE for dialogues than the best models achieve on the CNN/DM news dataset. 

According to ROUGE metrics, the best performing model is {\it DynamicConv} with {\it GPT-2} embeddings, trained on joined news and dialogue data with an utterance separation token.

\begin{table}[t!]
\begin{center}
\begin{tabular}{|l|c c c|}
\hline \bf Model & \bf  R-1 & \bf R-2 & \bf R-L \\ \hline
Lead-3 baseline & 40.24 & 17.44 & 34.90 \\
\hline
Pointer Generator & 38.72 & 16.67 & 35.59 \\
Fast Abs RL & \bf 40.99 & \bf 17.72 & \bf 38.30 \\
Transformer & 38.72 & 16.89 & 35.74 \\
LightConv & 39.44 & 17.20 & 36.20 \\
DynamicConv & 39.46 & 17.33 & 36.29 \\
\shortstack[l]{LightConv \\ \hspace{5mm}+ GPT2 emb} & 39.52 & 17.31 & 36.15 \\
\shortstack[l]{DynamicConv \\ \hspace{5mm}+ GPT2 emb} & 39.94 & 17.56 & 36.51 \\
\hline
\end{tabular}
\end{center}
\caption{Model evaluation on the news corpus test set}
\label{table:news-results}
\end{table}

\begin{table*}[t!]
\begin{center}
\begin{tabular}{|l|l|c|c c c|}
\hline \bf Model & \bf Train data & \bf Separator & \bf R-1 & \bf R-2 & \bf R-L \\ \hline
LONGEST-3 baseline & n/a & n/a & 32.46 & 10.27 & 29.92 \\
\hline
Pointer Generator & dialogues & no & 38.55 & 14.14 & 34.85 \\
Pointer Generator & dialogues & yes & 40.08 & 15.28 & 36.63 \\
Fast Abs RL & dialogues & no & 40.96 & 17.18 & 39.05 \\
Fast Abs RL Enhanced & dialogues & no & 41.95 & 18.06 & 39.23 \\
Transformer & dialogues & no & 36.62 & 11.18 & 33.06 \\
Transformer & dialogues & yes & 37.27 & 10.76 & 32.73 \\
LightConv & dialogues & no & 33.19 & 11.14 & 30.34 \\
DynamicConv & dialogues & no & 33.79 & 11.19 & 30.41 \\
DynamicConv & dialogues & yes & 33.69 & 10.88  & 30.93 \\ 
LightConv + GPT-2 emb. & dialogues & no & 41.81 & 16.34 & 37.63 \\
DynamicConv + GPT-2 emb. & dialogues & no & 41.79 & 16.44 & 37.54 \\
DynamicConv + GPT-2 emb. & dialogues & yes & 41.54 & 16.29 & 37.07 \\ 
\hline
Pointer Generator & news + dialogues & no & 35.04 & 13.25 & 32.42 \\
Pointer Generator & news + dialogues & yes & 37.27 & 14.42 & 34.36 \\
Fast Abs RL & news + dialogues & no & 41.03 & 16.93 & 39.05\\
Fast Abs RL Enhanced & news + dialogues & no & 41.87 & 17.47 & 39.53 \\
Transformer & news + dialogues & no & 41.91 & 18.25 & 38.77 \\
Transformer & news + dialogues & yes & 42.37 & 18.44 & 39.27 \\
LightConv & news + dialogues & no & 40.29 & 17.28 & 36.81 \\
DynamicConv & news + dialogues & no & 40.66 & 17.41 & 37.20 \\
DynamicConv & news + dialogues & yes & 41.07 & 17.11 & 37.27 \\ 
LightConv + GPT-2 emb. & news + dialogues & no & 44.47 & 19.75 & 40.07 \\
DynamicConv + GPT-2 emb. & news + dialogues & no & 44.69 & 20.28 & 40.76 \\
DynamicConv + GPT-2 emb. & news + dialogues & yes & \bf 45.41 & \bf 20.65 & \bf 41.45 \\
\hline
\end{tabular}
\end{center}
\caption{Model evaluation on the dialogues corpus test set}
\label{table:dialogues-results}
\end{table*}

\section{Linguistic verification of summaries} \label{Linguistic}
ROUGE is a~standard way of evaluating the quality of machine generated summaries by comparing them with reference ones. 
The metric based on \mbox{n-gram} overlapping, however, may not be very informative for abstractive summarization, where paraphrasing is a~keypoint in producing high-quality sentences.
To quantify this conjecture, we manually evaluated summaries generated by the models for 150 news and 100 dialogues. We asked two linguists to mark the quality of every summary on the scale of $-1$, $0$, $1$, where $-1$ means that a summarization is poor, extracts irrelevant information or does not make sense at all, $1$ -- it is understandable and gives a brief overview of the text, and $0$~stands for a summarization that extracts only a~part of relevant information, or makes some mistakes in the produced summary. 

We noticed a few annotations (7 for news and 4 for dialogues) with opposite marks (i.e. one annotator judgement was $-1$, whereas the second one was $1$) and decided to have them annotated once again by another annotator who had to resolve conflicts. For the rest, we calculated the linear weighted Cohen's kappa coefficient~\cite{cohen-kappa} between annotators' scores.  For news examples, we obtained agreement on the level of $0.371$ and for dialogues -- $0.506$. The annotators' agreement is higher on dialogues than on news, probably because of structures of those data -- articles are often long and it is difficult to decide what the key-point of the text is; dialogues, on the contrary, are rather short and focused mainly on one topic.

For manually evaluated samples, we calculated ROUGE metrics and the mean of two human ratings; the prepared statistics is presented in Table~\ref{table:human-evaluation}. As we can see, models generating dialogue summaries can obtain high ROUGE results, but their outputs are marked as poor by human annotators. Our conclusion is that the ROUGE metric corresponds with the quality of generated summaries for news much better than for dialogues, confirmed by Pearson's correlation between human evaluation and the ROUGE metric, shown in~Table~\ref{table:pearson-correlation}.

\begin{table*}[t!]
\begin{center}
\begin{tabular}{|l|l|c c c c c c|}
\hline \multicolumn{2}{|c|}{} & \bf \#examples & \bf mean & \bf median & \bf R-1 & \bf R-2 & \bf R-L \\ \hline
\multirow{3}{*}{NEWS}   & overall           & 100 & 0.18 & 0.5 & 39.76 & 16.55 & 36.23 \\ 
                        & Fast Abs RL       & 50  & 0.33 & 0.5 & 42.33 & 18.28 & 38.82 \\
                        & DynamicConv       & 50  & 0.03 & 0.25 & 37.19 & 14.81 & 33.64 \\
\hline
\multirow{4}{*}{DIALOGUES}  & overall               & 150 & -0.503  & -0.5  & 43.53 & 19.94 & 40.66 \\ 
                            & Fast Abs RL           & 50  & -0.55   & -0.75 & 42.16 & 19.28 & 40.37 \\
                            & Fast Abs RL Enhanced  & 50  & -0.63   & -1.0  & 39.79 & 16.59 & 37.05 \\
                            &  DynamicConv & \multirow{2}{*}{50}  & \multirow{2}{*}{-0.33}   & \multirow{2}{*}{-0.5}  & \multirow{2}{*}{48.63} & \multirow{2}{*}{23.95} & \multirow{2}{*}{44.57} \\
                            & \hspace{5mm} + GPT-2 emb. & & & & & & \\
\hline
\end{tabular}
\end{center}
\caption{Statistics of human evaluation of summaries' quality and ROUGE evaluation of those summaries}
\label{table:human-evaluation}
\end{table*}

\begin{table*}[t]
\begin{center}
\begin{tabular}{|l|c c|c c|c c|}
\hline \multirow{2}{*}{} & \multicolumn{2}{|c|}{\bf ROUGE-1} & \multicolumn{2}{|c|}{\bf ROUGE-2} & \multicolumn{2}{|c|}{\bf ROUGE-L} \\ \cline{2-7}
& \bf corr & \bf p-value & \bf corr & \bf p-value & \bf corr & \bf p-value \\ \hline
NEWS            & 0.47 & 1e-6 & 0.44 & 6e-6 & 0.48 & 1e-6 \\ 
\hline
DIALOGUES       & 0.32 & 7.7e-5 & 0.30 & 1.84e-4 & 0.32 & 8.1e-5 \\ 
\hline
\end{tabular}
\end{center}
\caption{Pearson's correlations between human judgement and ROUGE metric}
\label{table:pearson-correlation}
\end{table*}


\section{Difficulties in dialogue summarization} \label{Difficulties}
In a structured text, such as a news article, the information flow is very clear. However, in a~dialogue, which contains discussions (e.g. when people try to agree on a date of a meeting), questions (one person asks about something and the answer may appear a few utterances later) and greetings, most important pieces of information are scattered across the utterances of different speakers. What is more, articles are written in the third-person point of view, but in a~chat everyone talks about themselves, using a variety of pronouns, which further complicates the structure. Additionally, people talking on messengers often are in a hurry, so they shorten words, use the slang phrases (e.g. 'u r gr8' means 'you are great') and make typos. These phenomena increase the difficulty of performing dialogue summarization. 

Table \ref{tab:1_2_dials} and \ref{tab:all_dials} show a~few selected dialogues, together with summaries produced by the best tested~models:
\begin{itemize}
    \item {\it DynamicConv} + {\it GPT-2} embeddings with a~separator (trained on news + dialogues),
    \item {\it DynamicConv} + {\it GPT-2} embeddings (trained on news + dialogues),
    \item {\it Fast Abs RL} (trained on dialogues),
    \item {\it Fast Abs RL Enhanced} (trained on dialogues),
    \item {\it Transformer} (trained on news + dialogues).
\end{itemize}

One can easily notice problematic issues. Firstly, the models frequently have difficulties in associating names with actions, often repeating the same name, e.g., for Dialogue 1 in Table \ref{tab:1_2_dials}, {\it Fast Abs RL} generates the following summary: 'lilly and lilly are going to eat salmon'. To help the model deal with names, the utterances are enhanced by adding information about the other interlocutors -- {\it Fast Abs RL enhanced} variant described in Section \ref{models-params}. In this case, after enhancement, the model generates a summary containing both interlocutors' names: 'lily and gabriel are going to pasta...'. 
Sometimes models correctly choose speakers' names when generating a~summary, but make a~mistake in deciding who performs the action (the subject) and who receives the action (the object), e.g. for Dialogue 4 {\it DynamicConv + GPT-2 emb. w/o sep.} model generates the summary 'randolph will buy some earplugs for maya', while the correct form is 'maya will buy some earplugs for randolph'. 

A~closely related problem is capturing the context and extracting information about the arrangements after the discussion. 
For instance, for Dialogue 4, the {\it Fast Abs RL} model draws a wrong conclusion from the agreed arrangement. 
This issue is quite frequently visible in summaries generated by {\it Fast Abs RL}, which may be the consequence of the way it is constructed; it first chooses important utterances, and then summarizes each of them separately. This leads to the narrowing of the context and loosing important pieces of information.

One more aspect of summary generation is deciding which information in the dialogue content is important. For instance, for Dialogue 3 
{\it DynamicConv + GPT-2 emb. with sep.} generates a correct summary, but focuses on a~piece of information different than the one included in the reference summary.
In contrast, some other models -- like {\it Fast Abs RL enhanced} -- select both of the pieces of information appearing in the discussion. On the other hand, when summarizing Dialogue~5, the models seem to focus too much on the phrase 'it's the best place', intuitively not the most important one to summarize. 

\begin{table*}[ht!]
    \centering
    \begin{tabular}{|l|l|} 
        \hline
        \bf Dialogue 1 & \bf Dialogue 2 \\ 
        1. \textcolor{lasallegreen}{lilly:} sorry, i'm gonna be late & 
        1. \textcolor{lasallegreen}{randolph:} honey \\
        2. \textcolor{lasallegreen}{lilly:} don't wait for me and order the food &
        2. \textcolor{lasallegreen}{randolph:} are you still in the pharmacy? \\
        3. \textcolor{raspberry}{gabriel:} no problem, shall we also order & 
        3. \textcolor{raspberry}{maya:} yes  \\
        something for you? & 
        4. \textcolor{lasallegreen}{randolph:} buy me some earplugs please \\
        4. \textcolor{raspberry}{gabriel:} so that you get it as soon as you get & 
        5. \textcolor{raspberry}{maya:} how many pairs? \\
        to us? & 
        6. \textcolor{lasallegreen}{randolph:} 4 or 5 packs \\
        5. \textcolor{lasallegreen}{lilly:} good idea &  
        7. \textcolor{raspberry}{maya:} i'll get you 5 \\
        6. \textcolor{lasallegreen}{lilly:} pasta with salmon and basil is always  &
        8. \textcolor{lasallegreen}{randolph:} thanks darling \\ 
        very tasty here & \\
        \hline
        {\bf REF:} lilly will be late. gabriel will order pasta &
        {\bf REF:} maya will buy 5 packs of earplugs for \\
        with salmon and basil for her. &
        randolph at the pharmacy. \\
        & \\
        {\bf L3:} 6, 3, 4 \textcolor{frenchblue}{[38/17/38]} & 
        {\bf L3:} 2, 4, 8 \textcolor{frenchblue}{[36/8/36]} \\
        {\bf DS:} lilly and gabriel are going to order pasta &  
        {\bf DS:} randolph and maya are going to buy some \\
        with salmon and basil \textcolor{frenchblue}{[62/42/62]} &
        earplugs for randolph. \textcolor{frenchblue}{[43/19/43]} \\
        {\bf D:} lilly and gabriel are going to order pasta & {\bf D:} randolph will buy some earplugs for maya. \\
        with salmon and basil \textcolor{frenchblue}{[62/42/62]} & \textcolor{frenchblue}{[63/24/42]}\\
        {\bf F:} lilly will be late . she will order the food .  lilly & {\bf F:} maya is in the pharmacy . maya will get 5 . \\
        and lilly are going to eat salmon and basil &  \textcolor{frenchblue}{[48/21/48]} \\
        \textcolor{frenchblue}{[55/39/55]} & \\
        {\bf FE:} lilly will be late . lilly and gabriel are going & {\bf FE}: randolph is in the pharmacy . randolph \\
         to pasta with salmon and basil is always tasty . & will buy some earplugs for randolph . maya will \\
      \textcolor{frenchblue}{[63/47/63]} & get 5 .  \textcolor{frenchblue}{[64/38/64]} \\
        {\bf T:} lilly will order the food as soon as she gets to & {\bf T:} randolph will buy some earplugs for  \\
        gabriel \textcolor{frenchblue}{[31/17/23]} & randolph . maya will get 5 pairs . \textcolor{frenchblue}{[58/36/42]} \\
        \hline
    \end{tabular}
    \caption{Examples of dialogues (Part 1). REF -- reference summary, L3 -- LONGEST-3 baseline, DS -- DynamicConv + GPT-2 emb. with sep., D -- DynamicConv + GPT-2 emb., F -- Fast Abs RL, FE -- Fast Abs RL Enhanced, T --  Transformer. For L3, three longest utterances are listed. Rounded ROUGE values \textcolor{frenchblue}{[R-1/R-2/R-L]} are given in square brackets.}
    \label{tab:1_2_dials}
\end{table*}

\begin{table*}[ht!]
    \centering
    \begin{tabular}{|l|l|} 
        \hline
        \bf Dialogue 3 & \bf Dialogue 4 \\ 
        1. \textcolor{lasallegreen}{ashleigh:} looks like we're going to the cinema!!  & 1. \textcolor{lasallegreen}{paul:} what color flowers should i get \\
        2. \textcolor{lasallegreen}{ashleigh:} $<$file\_gif$>$ & 2. \textcolor{raspberry}{cindy:} any just not yellow \\ 
        3. \textcolor{raspberry}{peter:} you got the job?? & 3. \textcolor{lasallegreen}{paul:} ok, pink? \\
        4. \textcolor{lasallegreen}{ashleigh:} i got hte job! :d & 4. \textcolor{raspberry}{cindy:} no maybe red \\ 
        5. \textcolor{raspberry}{peter:} $<$file\_gif$>$ & 5. \textcolor{lasallegreen}{paul:} just tell me what color and what type \\
        6. \textcolor{lasallegreen}{ashleigh:} $<$file\_gif$>$ & ok? \\
         & 6. \textcolor{raspberry}{cindy:} ugh, red roses! \\
        \hline
        {\bf REF:} ashleigh got the job. & {\bf REF:} paul will buy red roses following cindy's \\
					 & advice. \\ & \\
		{\bf L3:} 1, 4, 3 \textcolor{frenchblue}{[33/18/33]} & {\bf L3:} 5, 1, 2 \textcolor{frenchblue}{[13/0/13]} \\
        {\bf DS:} ashleigh and peter are going to the cinema. & {\bf DS:} paul and cindy don't like red roses.\\
	 \textcolor{frenchblue}{[33/0/33]} &  \textcolor{frenchblue}{[47/13/35]} \\ 
        {\bf D:} ashleigh got hte job. \textcolor{frenchblue}{[75/33/75]} & {\bf D:} paul asks cindy what color flowers should \\
         &  buy.  \textcolor{frenchblue}{[35/0/24]} \\ 
         {\bf F:} ashleigh and ashleigh are going to the cinema. & {\bf F:} cindy is going to buy red roses \textcolor{frenchblue}{[50/29/38]} \\
         peter got the job . \textcolor{frenchblue}{[50/29/50]}  & \\
        {\bf FE:} ashley and peter are going to the cinema & {\bf FE:} cindy is buying red roses . cindy will buy \\
	 together . ashleigh got the job . \textcolor{frenchblue}{[47/40/47]} & red .  \textcolor{frenchblue}{[56/38/44]} \\
        {\bf T:} ashleigh got the job at the cinema . peter and & {\bf T:} cindy does n't know what color should get. \\
	 ashleigh are going there . \textcolor{frenchblue}{[47/40/47]} & cindy does not know what to do \textcolor{frenchblue}{[8/0/8]} \\ \hline 
	 \hline
	 \multicolumn{2}{|l|}{\bf Dialogue 5} \\
	 \multicolumn{2}{|l|}{1. \textcolor{lasallegreen}{eve:} where are we meeting?} \\
        \multicolumn{2}{|l|}{2. \textcolor{raspberry}{charlie:} at the entrance} \\
        \multicolumn{2}{|l|}{3. \textcolor{goldenpoppy}{nicole:} yes, it's the best place. we would't find each other inside, it'll be too crowded} \\
        \multicolumn{2}{|l|}{4. \textcolor{lasallegreen}{eve:} ok!} \\ 
        \hline
        \multicolumn{2}{|l|}{{\bf REF:} eve, charlie and nicole are meeting at the entrance. }\\
        \multicolumn{2}{|l|}{} \\
        \multicolumn{2}{|l|}{{\bf L3:} 3, 1, 2 \textcolor{frenchblue}{[43/11/43]}} \\
                \multicolumn{2}{|l|}{{\bf DS:} eve, charlie and nicole are meeting at the entrance. \textcolor{frenchblue}{[100/100/100]}}\\
        \multicolumn{2}{|l|}{{\bf D:} eve, charlie and nicole are meeting at the entrance. \textcolor{frenchblue}{[100/100/100]}} \\
        \multicolumn{2}{|l|}{{\bf F:} charlie is at the entrance . it 's the best place . \textcolor{frenchblue}{[42/24/42]}} \\
        \multicolumn{2}{|l|}{{\bf FE:} charlie is at the entrance . nicole and charlie are going to find each other inside . \textcolor{frenchblue}{[58/18/42]}} \\
        \multicolumn{2}{|l|}{{\bf T:} eve and nicole are meeting at the entrance . it 's the best place to meet . \textcolor{frenchblue}{[67/55/67]}} \\
        \hline
    \end{tabular}
    \caption{Examples of dialogues (Part 2). REF -- reference summary, L3 -- LONGEST-3 baseline, DS -- DynamicConv + GPT-2 emb. with sep., D -- DynamicConv + GPT-2 emb., F -- Fast Abs RL, FE -- Fast Abs RL Enhanced, T --  Transformer. For L3, three longest utterances are listed. Rounded ROUGE values \textcolor{frenchblue}{[R-1/R-2/R-L]} are given in square brackets.}
    \label{tab:all_dials}
\end{table*}

\section{Discussion} \label{Discussion}
This paper is a step towards abstractive summarization of dialogues by (1) introducing a new dataset, created for this task, (2) comparison with news summarization by the means of automated (ROUGE) and human evaluation. 

Most of the tools and the metrics measuring the quality of text summarization have been developed for a~single-speaker document, such as news; as such, they are not necessarily the best choice for conversations with several speakers.

We test a few general-purpose summarization models. In terms of human evaluation, the results of dialogues summarization are worse than the results of news summarization. This is connected with the fact that the dialogue structure is more complex -- information is spread in multiple utterances, discussions, questions, more typos and slang words appear there, posing new challenges for summarization. On the other hand, dialogues are divided into utterances, and for each utterance its author is assigned. We demonstrate in experiments that the models benefit from the introduction of separators, which mark utterances for each person. This suggests that dedicated models having some architectural changes, taking into account the assignation of a~person to an utterance in a systematic manner, could improve the quality of dialogue summarization.

We show that the most popular summarization metric ROUGE does not reflect the quality of a~summary. Looking at the ROUGE scores, one concludes that the dialogue summarization models perform better than the ones for news summarization. In fact, this hypothesis is not true -- we performed an independent, manual analysis of summaries and we demonstrated that high ROUGE results, obtained for automatically-generated dialogue summaries, correspond with lower evaluation marks given by human annotators. An interesting example of the misleading behavior of the ROUGE metrics is presented in Table \ref{tab:all_dials} for Dialogue 4, where a wrong summary -- 'paul and cindy don't like red roses.' -- obtained all ROUGE values higher than a correct summary -- 'paul asks cindy what color flowers should buy.'.  
Despite lower ROUGE values, news summaries were scored higher by human evaluators.
We conclude that when measuring the quality of model-generated summaries, the ROUGE metrics are more indicative for news than for dialogues, and a new metric should be designed to measure the quality of abstractive dialogue summaries.

\section{Conclusions} \label{Conclusion}

In our paper we have studied the challenges of abstractive dialogue summarization. We have addressed a major factor that prevents researchers from engaging into this problem: the lack of a~proper dataset. 
To the best of our knowledge, this is the first attempt to create a~comprehensive resource of this type which can be used in future research. The next step could be creating an even more challenging dataset with longer dialogues that not only cover one topic, but span over numerous different ones. 

As shown, summarization of dialogues is much more challenging than of news. In order to perform well, it may require designing dedicated tools, but also new, non-standard measures to capture the quality of abstractive dialogue summaries in a relevant way. We hope to tackle these issues in future work.
\vfill

\section*{Acknowledgments}
We would like to express our sincere thanks to Tunia B\l{}achno, Oliwia Ebebenge, Monika J{\k e}dras and Ma\l{}gorzata Krawentek for their huge contribution to the corpus collection -- without their ideas, management of the  linguistic task and verification of examples we would not be able to create this paper. We are also grateful for the reviewers' helpful comments and suggestions.





\clearpage
\bibliography{emnlp-ijcnlp-2019}
\bibliographystyle{acl_natbib}

\end{document}